\documentclass[10pt,twocolumn,letterpaper]{article}

\usepackage[pagenumbers]{cvpr} 

\usepackage[dvipsnames]{xcolor}

\usepackage{multirow}
\usepackage{tabularx}
\usepackage[vlines]{tabularht}
\usepackage{paralist}
\usepackage[linesnumbered,ruled,lined,boxed]{algorithm2e}

\SetCommentSty{mycommfont}
\SetKwInput{kwHyper}{Hyperparameters}

\definecolor{cvprblue}{rgb}{0.21,0.49,0.74}
\usepackage[pagebackref,breaklinks,colorlinks,citecolor=cvprblue,linkcolor=cvprblue,urlcolor=cvprlbue]{hyperref}
\usepackage{subcaption}
\def\paperID{5845} 
\def\confName{CVPR}
\def\confYear{2024}

\title{VidToMe: Video Token Merging for Zero-Shot Video Editing}

\hypersetup{
    urlcolor=cvprblue,
}
\author{Xirui Li$^{1}$ \hspace{4mm}Chao Ma$^{1}$\hspace{4mm}Xiaokang Yang$^{1}$\hspace{4mm}Ming-Hsuan Yang$^{2}$\\
$^{1}$Shanghai Jiao Tong University\hspace{4mm}$^{2}$UC Merced
\\
{\small Project webpage: \url{https://vidtome-diffusion.github.io}}
\vspace{-2.5mm}
}

\begin{document}
\maketitle
\begin{abstract}
Diffusion models have made significant advances in generating high-quality images, but their application to video generation has remained challenging due to the complexity of temporal motion. Zero-shot video editing offers a solution by utilizing pre-trained image diffusion models to translate source videos into new ones. Nevertheless, existing methods struggle to maintain strict temporal consistency and efficient memory consumption. In this work, we propose a novel approach to enhance temporal consistency in generated videos by merging self-attention tokens across frames. By aligning and compressing temporally redundant tokens across frames, our method improves temporal coherence and reduces memory consumption in self-attention computations. The merging strategy matches and aligns tokens according to the temporal correspondence between frames, facilitating natural temporal consistency in generated video frames. To manage the complexity of video processing, we divide videos into chunks and develop intra-chunk local token merging and inter-chunk global token merging, ensuring both short-term video continuity and long-term content consistency. Our video editing approach seamlessly extends the advancements in image editing to video editing, rendering favorable results in temporal consistency over state-of-the-art methods.
\end{abstract}    
\section{Introduction}
\label{sec:intro}
\begin{figure}[t]
    \centering
    \includegraphics[width=\linewidth]{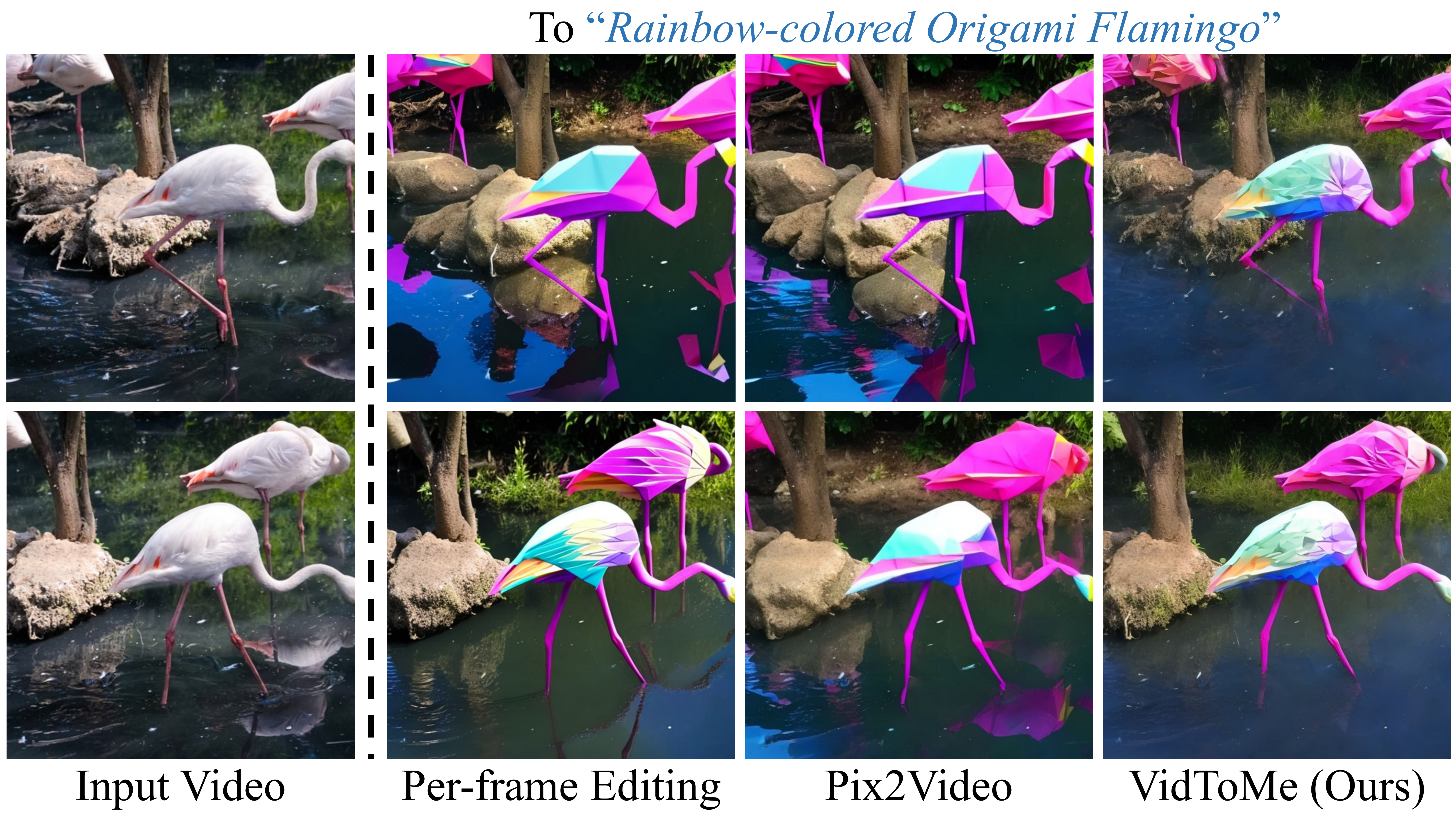}
    \caption{Given an input source video and a text prompt, we leverage a pre-trained image diffusion model~\cite{rombach2022high} to edit the video.
    The state-of-the-art zero-shot video editing approaches, e.g., Pix2Video~\cite{ceylan2023pix2video}, struggle to generate temporal consistent frames with self-attention extension.
    Our proposed method merges tokens across frames, rendering higher temporal consistency.}
    \label{fig:sa}
\end{figure}

Diffusion models~\cite{ho2020denoising, song2020score, dhariwal2021diffusion, croitoru2023diffusion, rombach2022high, song2020denoising} have made significant advances in synthesizing media content, allowing for the creation of diverse, high-quality images. However, diffusion models have yet to achieve high quality in generating videos. Due to the complexity of temporal motion in videos, training a video diffusion model requires a massive amount of data and computation resources. 
To avoid learning temporal motion from scratch, zero-shot video editing leverages a pre-trained image diffusion model to translate a source video into a new one, retaining motion from the source video. Separately editing each frame likely results in inconsistent frames (Fig~\ref{fig:sa} Per-frame Editing).
Existing video editing methods~\cite{wu2023tune, ceylan2023pix2video, qi2023fatezero, wang2023zero, yang2023rerender} typically extend the self-attention modules of diffusion models to process multiple frames jointly instead of separately. Despite the promise, two issues ensue with such approaches.
First, though cross-frame attention encourages a roughly consistent appearance, the generated frames lack strict consistency in details. As human perception is sensitive to video continuity, tiny changes or jittering between frames can significantly degrade the quality of generated videos. Second, including multi-frame tokens in self-attention increases memory consumption quadratically. 
Computing self-attention on four frames requires 16 times larger GPU memory than on one frame. 
With these limitations, state-of-the-art video editing methods such as Pix2Video~\cite{ceylan2023pix2video} struggle to generate temporal consistent videos, as shown in Fig.~\ref{fig:sa}. 
Thus, it is imperative to develop effective and efficient diffusion-based zero-shot video editing methods.

\begin{figure}[t]
    \centering
    \includegraphics[width=1\linewidth]{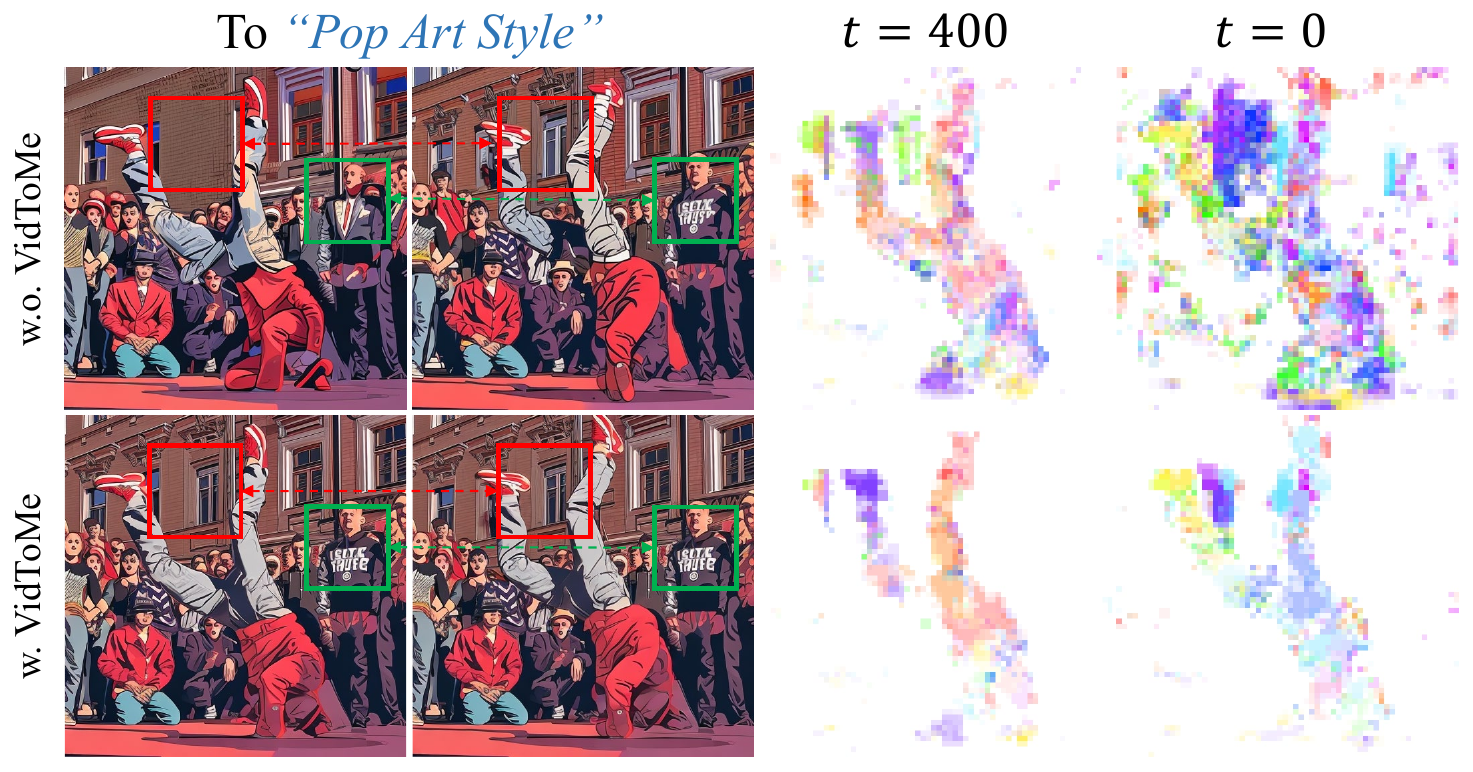}
    \caption{Comparison of frames edited to \textit{"Pop Art Style"} by PnP~\cite{tumanyan2023plug} with or without VidToMe. Left: Edit results. Right: Visualized token matching between two frames as flow maps. Color represents the direction of the matched token in another frame. We label the denoising timestamp above ($1000\rightarrow 0$). Our method aligns correspondent tokens and fixes the inconsistencies (window, clothes) in per-frame editing.}
    \label{fig:visflow}
\end{figure}

In this work, we present a novel method, VidToMe, to enhance the temporal consistency of generated videos by merging the tokens in diffusion models across video frames.
Our motivation comes from the recent developments~\cite{bolya2022tome,bolya2023tomesd} in compressing transformer tokens to improve computational efficiency. 
For video editing, we observe that the transformer tokens of video frames are much more correlated in the temporal domain than in the spatial domain. 
We can align tokens across frames according to the correlation and compress the temporally redundant tokens to facilitate joint self-attention.
We show that when multiple frames share the same set of tokens in the self-attention module, the diffusion model naturally generates temporally consistent video frames.
Hence, we propose merging tokens over time to compress and unify the internal diffusion feature embeddings, achieving temporal consistency in generated videos and reducing memory consumption in computing self-attention across frames.

Our method involves merging tokens in one frame with the most similar ones in another frame. 
As shown in Fig.~\ref{fig:visflow}, this merging strategy allows us to match and align tokens according to the temporal correspondence between frames.
Thus, applying VidToMe fixes the misalignment of details in per-frame editing.
As processing all frames at once is not feasible, we divide the video into chunks and perform intra-chunk local token merging and inter-chunk global token merging, ensuring both short-term and long-term consistency. 
Short-term consistency improves video continuity, while long-term consistency prevents the video content from drifting over time. 
Note that our video editing method can be seamlessly integrated with existing controlling schemes~\cite{zhang2023adding,tumanyan2023plug}, taking full advantage of the advancements in image editing for video editing.
Extensive experiments show that the proposed video editing method performs well regarding temporal consistency and text alignment over the state-of-the-art approaches.
The main contributions of this work are three-fold:
\begin{compactitem}
    \item We propose a novel diffusion-based approach, VidToMe, to merge self-attention tokens across frames when generating video frames, improving temporal consistency and computational efficiency.
    \item We design a video editing pipeline that jointly generates all video frames with short-term local token merging and long-term global token merging to enforce feature alignment throughout the video.
    \item We comprehensively evaluate our method to show the 
    state-of-the-art video editing performance.
\end{compactitem}

\section{Related Work}
\label{sec:related}

As several thorough surveys on image and video generation~\cite{croitoru2023diffusion, creswell2018generative, aldausari2022video,liu2021generative} exist in the literature, here we discuss diffusion models and related image and video editing schemes.

\noindent \textbf{Diffusion-based Image and Video Synthesis.}
Diffusion Models (DM)~\cite{sohl2015deep,ho2020denoising, song2020score} have recently achieved state-of-the-art performance in numerous tasks, including image generation~\cite{dhariwal2021diffusion, nichol2021improved, croitoru2023diffusion, rombach2022high, song2020denoising}.
DMs learn to reverse a forward diffusion process and generate an image by gradually denoising it from pure noise.
Notable examples of improving DMs include~\cite{karras2022elucidating, ho2021classifier, salimans2021progressive} and numerous applications~\cite{avrahami2022blended, mokady2023null, gal2022image, kawar2022denoising, li2022srdiff, lugmayr2022repaint, meng2021sdedit, ruiz2023dreambooth}. Benefiting from large-scale pretraining~\cite{radford2021learning, schuhmann2022laion}, text-to-image DMs have shown impressive results in generating high-quality and diverse images~\cite{rombach2022high, saharia2022photorealistic, ramesh2022hierarchical, nichol2022glide}.
Naturally, DMs have been applied to video synthesis, typically by incorporating temporal layers into image DMs~\cite{blattmann2023align,ho2022imagen,ho2022video}.
Despite the demonstrated success in unconditional video generation~\cite{ho2022video,yu2023video}, text-to-video DMs are not as satisfying as image ones. 
Due to the complexity of temporal motion, training video DMs requires intensive computation resources and large-scale annotated video datasets, which significantly hinders the progress of this field. 

\noindent \textbf{Diffusion-based Image Editing.}
In addition to text, some works have introduced additional control signals for image editing or controllable image generation~\cite{zhang2023adding,saharia2022palette,tumanyan2023plug}.
Some schemes introduce adapter layers~\cite{mou2023t2i} or other trainable modules ~\cite{voynov2023sketch,zhang2023adding} to accept additional control signals. 
ControlNet~\cite{zhang2023adding} supports various conditions such as edge maps, depth maps, and key points by finetuning an attached copy of DM. 
Other methods edit a source image by manipulating intermediate diffusion features~\cite{hertz2022prompt, tumanyan2023plug} or optimization-based guidance~\cite{epstein2023diffusion,parmar2023zero}.
Plug-and-Play~\cite{tumanyan2023plug} maintains image structure by injecting self-attention maps and internal features from the source image.
Self-guidance~\cite{epstein2023diffusion} and pix2pix-zero~\cite{parmar2023zero} edit the image by imposing a guidance loss optimized during generation.
StableDiffusion2~\cite{rombach2022high} presents a depth-conditioned model that directly includes the depth map in its input.
In this paper, we perform video editing by applying these image editing methods to video frames while preserving temporal coherence via merging video tokens. 

\noindent \textbf{Diffusion-based Video Editing.}
With the recent success of text-to-image DMs in powering text-driven image editing~\cite{tumanyan2023plug,hertz2022prompt,meng2021sdedit}, many works apply a pre-trained text-to-image DM~\cite{rombach2022high} for text-driven video editing. 
The critical problem is how to keep temporal coherency in generation.
Tune-A-Video~\cite{wu2023tune} inflates the DM with temporal attention layers and finetunes on the source video.
vid2vid-zero~\cite{wang2023zero} maintains the video structure by injecting cross-attention maps from the source video.
In~\cite{ceylan2023pix2video}, Pix2Video guides the generation with a reference frame by self-attention features injection and latent update.
Rerender-A-Video~\cite{yang2023rerender} fuses the previous frame warped by the source video optical flow and applies multi-stage latent operations.
On the other hand, Fate/Zero~\cite{qi2023fatezero} uses a dedicated attention blending block to inject attention maps from the source video.
TokenFlow~\cite{geyer2023tokenflow} shares a similar idea with our method to enforce temporal consistency by unifying self-attention tokens. It computes the inter-frame correspondences by extracting tokens from the source video. Then the tokens are propagated between the jointly-edited keyframes according to the correspondances.
Note that these methods commonly extend the self-attention modules into the spatial-temporal domain to encourage consistent appearance across frames.
However, extending self-attention does not enforce temporal consistency well and increases memory overhead. 
Our method simultaneously addresses these two problems by merging similar tokens across video frames. 

\section{Preliminaries}
\label{sec:prelim}
\noindent \textbf{Latent Diffusion Model.}
Diffusion models~\cite{sohl2015deep,ho2020denoising, song2020score} are a class of generative models based on an iterative denoising process.
An image DM supposes a forward process where a clean image $x_0$ is corrupted by Gaussian noise $\epsilon$,
\begin{equation}
    x_t = \sqrt{\alpha_t} x_0 + \sqrt{1-\alpha_t} \epsilon,
\end{equation}
where $t=1,\dots,T$ is the current timestep and $\{\alpha_t\}$ are the monotonically decreasing noise schedule.
Then, starting from random Gaussian noise, DM reverses the forward process to generate an image by estimating the noise direction and progressively denoising it.

Recent large-scale diffusion models~\cite{rombach2022high, saharia2022photorealistic, ramesh2022hierarchical} operate in the latent space to improve performance and efficiency. 
These latent diffusion models train an autoencoder~\cite{Kingma2014} to map the image between pixel and latent space. Let $\mathcal{E}(\cdot)$ and $\mathcal{D(\cdot)}$ be the encoder and the decoder, where $\mathcal{E}(x)=z, \mathcal{D}(z)\approx x$. 
Both the training and inference are conducted in the latent space.
Typically, a UNet~\cite{ronneberger2015u} $\epsilon_{\theta}$ is trained to estimate the noise with the objective 
\begin{equation}
    \min_{\theta} E_{z, \epsilon\sim \mathcal{N}(0,I),t} \| \epsilon - \epsilon_{\theta} (z_t,t,c)\|,
\end{equation}
where $c$ is the text embedding in text-to-image DMs. 
In this work, we base our experiments on Stable Diffusion~\cite{rombach2022high}, a large-scale text-to-image latent diffusion model.

\noindent \textbf{Token Merging.}
Token Merging (ToMe)~\cite{bolya2022tome} is a method to increase the throughput of existing ViT models by gradually merging redundant tokens in the transformer blocks. 
It combines similar tokens to reduce the redundancy as well as the number of tokens, speeding up the computation. 
Our method leverages the lightweight bipartite soft matching algorithm of ToMe to merge tokens across video frames.

Given input tokens $T$, the algorithm first partitions the tokens into a source ($src$) and destination ($dst$) set and computes the pair-wise cosine similarity between the two sets.
Then, $src$ tokens are linked to their most similar token in $dst$.
Next, $r$ most similar edges are selected, and connected tokens are merged. 
Finally, all the tokens are concatenated as the output.
We use the $dst$ token as the merged token value instead of averaging the value of merged tokens, which produces better results in practice.
Our method divides merged tokens after self-attention to keep the token number unchanged.
Like~\cite{bolya2023tomesd}, we divide a token simply by assigning its value to all restored tokens, which means a token merged from two tokens will be separated into two identical tokens.
We define the token merging and unmerging operations, M and U:
\begin{equation}
\begin{aligned}
&E=\text{Match}(src,dst,r), \\
&T_m=\text{M}(T,E),T_u=\text{U}(T_m,E).
\end{aligned}
\label{equ:tome}
\end{equation}
Match$(\cdot)$ outputs the matching map $E$ with $r$ edges from $src$ to $dst$. M$(\cdot)$ and U$(\cdot)$ merge and unmerge tokens according to matching $E$. 

\section{Proposed Method}
\label{sec:method}

\begin{figure*}
    \centering
    \includegraphics[width=.85\linewidth]{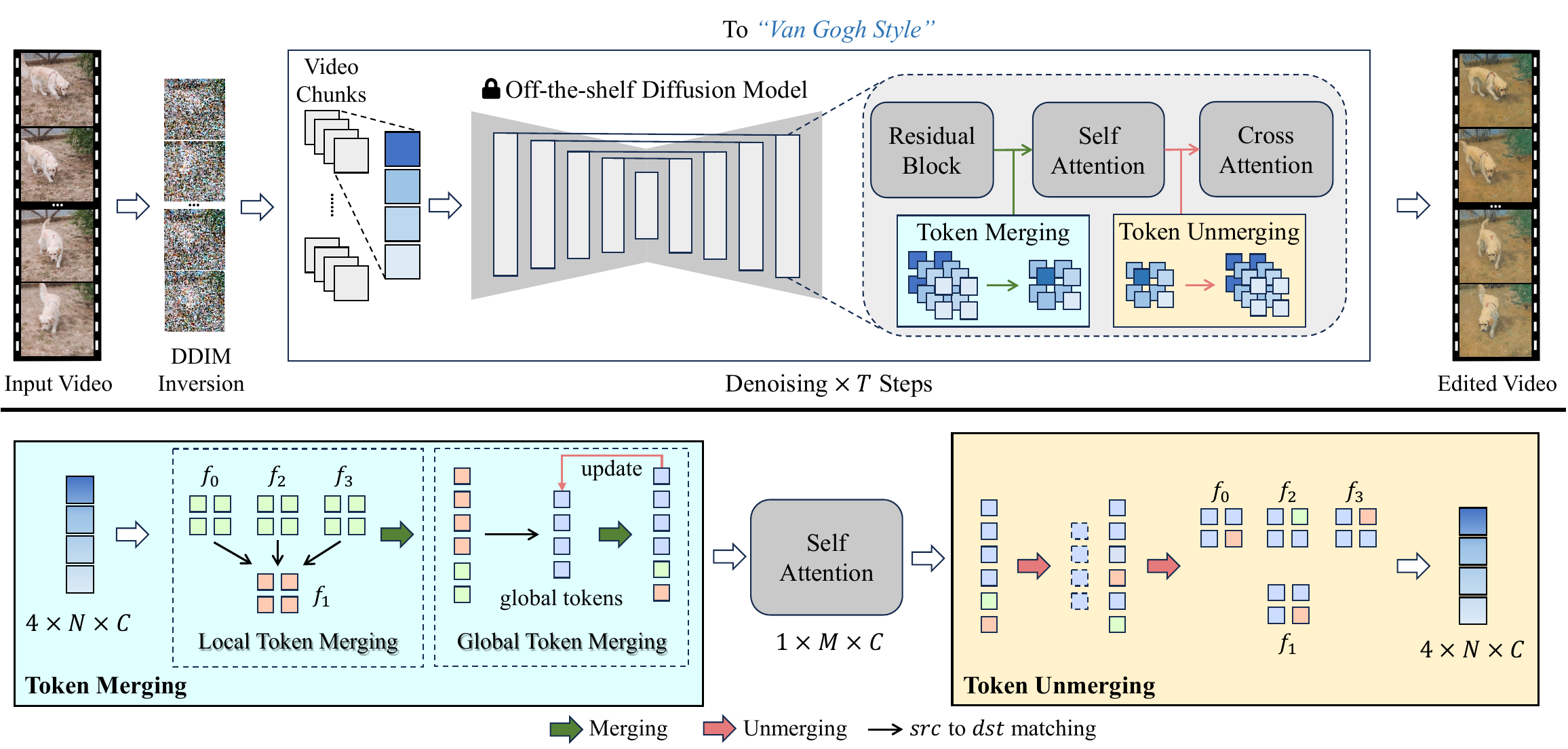}
    \caption{Pipeline of our proposed video editing method, VidToMe. We apply DDIM inversion on the source video frames to obtain the initial noisy latents. We denoise frame latents with an off-the-shelf text-to-image diffusion model, combining an existing controlling method~\cite{zhang2023adding, tumanyan2023plug}. In each iteration, frames are split into chunks and denoised by the diffusion model, where we attach our lightweight token merging and unmerging operation around the self-attention modules. We first merge tokens locally into a random frame in the chunk. Then, the merged tokens are combined with the global tokens maintained across chunks in one iteration. After self-attention, we unmerge the tokens to the original size for the following operations.}
    \label{fig:pipeline}
\end{figure*}

Our objective is to generate an edited video that matches a given editing prompt while preserving the motion and structure of a source video. 
To achieve this, we use a pre-trained text-to-image diffusion model to generate individual frames. 
We apply DDIM inversion~\cite{song2020denoising} and existing controlling methods~\cite{zhang2023adding, tumanyan2023plug, rombach2022high} for image editing to preserve the source frame structure.
However, more effort is required to achieve temporal consistency across video frames. 
%
%
We observe that transformer tokens are correlated across frames as the temporal correspondence in videos. 
Thus, we compress multi-frame tokens by merging similar tokens together so that the self-attention module extracts consistent features for each frame. 
The unified internal features promote the diffusion model to generate consistent video frames.
Fig.~\ref{fig:pipeline} presents an overview of the proposed method. 

The proposed video token merging strategy, VidToMe, is detailed at the bottom of Fig.~\ref{fig:pipeline}. 
We first merge tokens across frames in one video chunk to enforce short-term video continuity. 
The locally merged tokens are combined with a set of global tokens from previous chunks, enabling long-term token sharing. Joint self-attention extracts consistent features on merged tokens, which are then propagated to each frame by token unmerging.
Our video token merging algorithm has two advantages. 
First, merged tokens are shared across frames, enforcing temporal consistency. 
Second, token merging compresses redundant tokens in the self-attention module, improving computation efficiency. 
As a lightweight parameter-free operation, VidToMe can seamlessly integrate with existing image editing methods~\cite{zhang2023adding, tumanyan2023plug, rombach2022high}.
We first provide an overview of our video editing pipeline in Section~\ref{sec:pipeline} and then elaborate on our video token merging method in Section~\ref{sec:vidtome}.

\subsection{Video Editing Pipeline}\label{sec:pipeline}
Given the source video $\mathcal{V}$ with $n$ frames $(f_1,f_2,\dots,f_{n})$, we first invert each frame to noise by DDIM inversion. 
DDIM inversion applies the deterministic DDIM sampler~\cite{song2020denoising} in the reversed order $t=0\rightarrow T$, turning a clean latent $z_0$ into a noisy latent $z_T$. 
Using the inverted frames as the initial noise, we iteratively denoise them by an off-the-shelf text-to-image diffusion model with edit prompt $\mathcal{P}$ as the text condition.

Unlike existing methods~\cite{yang2023rerender,ceylan2023pix2video} that generate each frame separately, we process all frames together in each denoising iteration.
At iteration $t$, we randomly split frames $(z_t^{1},z_{t}^{2},\dots, z_{t}^{n})$ into a sequence of chunks $(C_1,C_2,\dots,C_m)$. 
Each chunk contains $B$ consecutive frames except for the first and last chunks.
The initial chunk $C_1=(z_t^{1},\dots,z_t^{b})$ where $b$ is a random integer in $[1,B]$.
The randomness here ensures the probability of frames split into a chunk proportional to the time interval, avoiding the ``chunks'' seen in the generated video. 
We process each chunk of frames by the diffusion model in a random order.

In the diffusion model, frames are concatenated in the batch dimension, which means the model treats all frames as separate images. 
To enforce temporal consistency, we merge tokens across frames in the self-attention module of the diffusion model. 
A diffusion model is typically realized as a UNet~\cite{ronneberger2015u}, composed of a series of downsample and upsample blocks. 
A layer in each block consists of a residual block, a self-attention module~\cite{vaswani2017attention}, and a cross-attention module. 
Among them, the self-attention module has been shown to be highly correlated to the structure and appearance of the image~\cite{tumanyan2023plug}. 
Thus, we merge multi-frame tokens before self-attention so that the self-attention jointly processes the merged tokens and outputs consistent features. 
Note that our method only changes the input tokens of the self-attention without any modification to the self-attention operation. 
To perform the following processes, we restore the output tokens to the original size by token unmerging.
As merging tokens in deep blocks may degrade the generation quality, we perform token merging in the first two downsample blocks and the last two upsample blocks.

After $T$ iterations of denoising, we obtain the edited frames latents $(z_0^{1},z_0^{2},\dots,z_0^{n})$ and the edited video $\mathcal{V}^{*}$ after VAE decoding. 
Our method works with an existing controlling method for image editing to preserve the structure of source frames, such as ControlNet~\cite{zhang2023adding}, Plug-and-Play (PnP)~\cite{tumanyan2023plug}, and depth-conditioned diffusion model~\cite{rombach2022high}.

\subsection{Video Token Merging} \label{sec:vidtome}
%
This section presents our video token merging strategy, which focuses on the self-attention module in the diffusion model.
A self-attention module takes a sequence of input tokens and outputs the same number of tokens.
The input and output tokens are denoted as $T_{in}$ and $T_{out}$, both belonging to the space $R^{B\times N\times C}$, where $B$ is the number of frames, $N$ is the number of tokens per frame, and $C$ is the feature dimension.
We enforce temporal coherence by merging input tokens across $B$ frames in one video chunk (local token merging) and merging global tokens from previous chunks of frames (global token merging).
After self-attention, we unmerge the output tokens to their original size.
These operations are performed on the input and output tokens without modifying the self-attention module. 

\noindent \textbf{Local Token Merging.}
Given a set of input tokens denoted by $T_{in}=\{T_{in}^{f}\}_{f=0}^{B-1}$,
we randomly select one out of $B$ frames as the current target frame, \eg, the $k^{th}$ frame.
We then apply the bipartite soft matching algorithm mentioned in Section~\ref{sec:prelim} and Equation~\ref{equ:tome} to merge the other frames to the target frame:
$$
T_{lm}=\text{M}(T_{in}, \text{Match}(T_{in}^{src}, T_{in}^{dst},r)), 
$$
where $T_{in}^{src}=\{T_{in}^{f}\}_{f=0,f\neq k}^{B-1}$ and $T_{in}^{dst}=T_{in}^{k}$.
We set $r=p(B-1)N$ where $(B-1)N$ is the $src$ token number and $p$ is the merging ratio. 
A large merging ratio (\eg, $p=0.9$) can be used as video frames are highly redundant. 
Local token merging enforces consistency in a small frame chunk.

However, for long-term consistency, we need more than short-term video continuity. 
For example, a video's first and last frames will never be processed in one chunk, leading to appearance drifting along the video.
Thus, we need another way to share tokens across the whole video. 
Enlarging the chunk size or implementing a hierarchical merging is helpful but requires an even larger memory capacity. 
Instead, we propose a simple yet effective global token merging strategy for long-term consistency.

\noindent \textbf{Global Token Merging.} 
At each iteration, we maintain a set of global tokens denoted as $T_g$ that spans across video chunks. 
The initial global tokens are set to be the locally merged tokens of the first chunk, \ie, $T_g^{1}=T_{lm}$. 
For the $k_{th}$ frame chunk, we merge its locally merged tokens $T_{lm}$ with the previous global tokens $T_g^{k-1}$ as the following operation:
\begin{equation}
    T_{gm}=\text{M}(\{T_{lm}, T_g^{k-1}\}, \text{Match}(T_{lm},T_{g}^{k-1},r)),
\end{equation}
where $T_{gm}$ represents the final input to the self-attention module. 
In practice, we randomly assign $src$ and $dst$ to local and global tokens.
We can update the global tokens to include the tokens from the current frames in several ways. 
One possible option is to use merged tokens $T_{gm}$ as new global tokens. 
However, this approach is not feasible for arbitrarily long videos as it always increases the number of global tokens. 
Instead, we unmerge $T_{gm}$ back to local tokens $T_{lm}^{u}$ and global tokens and set the current global tokens to be the unmerged local tokens, \ie, $T_{g}^{k}=T_{lm}^{u}$.

\noindent \textbf{Self-Attention Analysis.} 
We analyze the self-attention operation on merged tokens in more detail. 
The input, denoted as $T_{gm}$, comprises $M$ tokens.
We can infer $M=(0.11B+0.99)N=1.43N$, assuming chunk size $B=4$ and merging ratio $p=0.9$ for local and global merging. 
The tokens are mapped to $Q, K, V$ matrices during self-attention. 
For multiplication between $Q$ and $K$, the original input of size $4\times N\times C$ has a space complexity of $O(4N^2)$. 
In contrast, the complexity is reduced to half with the merged tokens $T_{gm}\in R^{1\times M\times C}$ as input, $O(M^2)\approx O(2N^{2})$. 
Essentially, our token merging method combines multiple frames into one, reducing redundancy among frames. Self-attention then identifies consistent features in this unified frame.

\noindent \textbf{Token Unmerging.}
The output tokens $T_o$ of the self-attention module need to be restored to their original shape as separate frames to perform the following image operations. 
As such, we first unmerge the tokens into local and global tokens and then unmerge the local tokens into $B$ separate frames, reversing the merging process. Denoting respective matching maps for local and global token merging as $E_l$ and $E_g$,
we formulate the token unmerging as $U(T_o, E_g)=(T_{local}, T_{global})$ to divide the local tokens and $U(T_{local}, E_l)=T_{out}$ to obtain the final output.
The unmerged tokens in the output are identical to the original merged tokens, ensuring consistency across frames.
\section{Experimental Results}
\label{sec:exp}

\begin{figure*}[t]
\begin{tabular}{@{}p{0.49\textwidth}p{0.49\textwidth}@{}}
  \includegraphics[width=\linewidth]{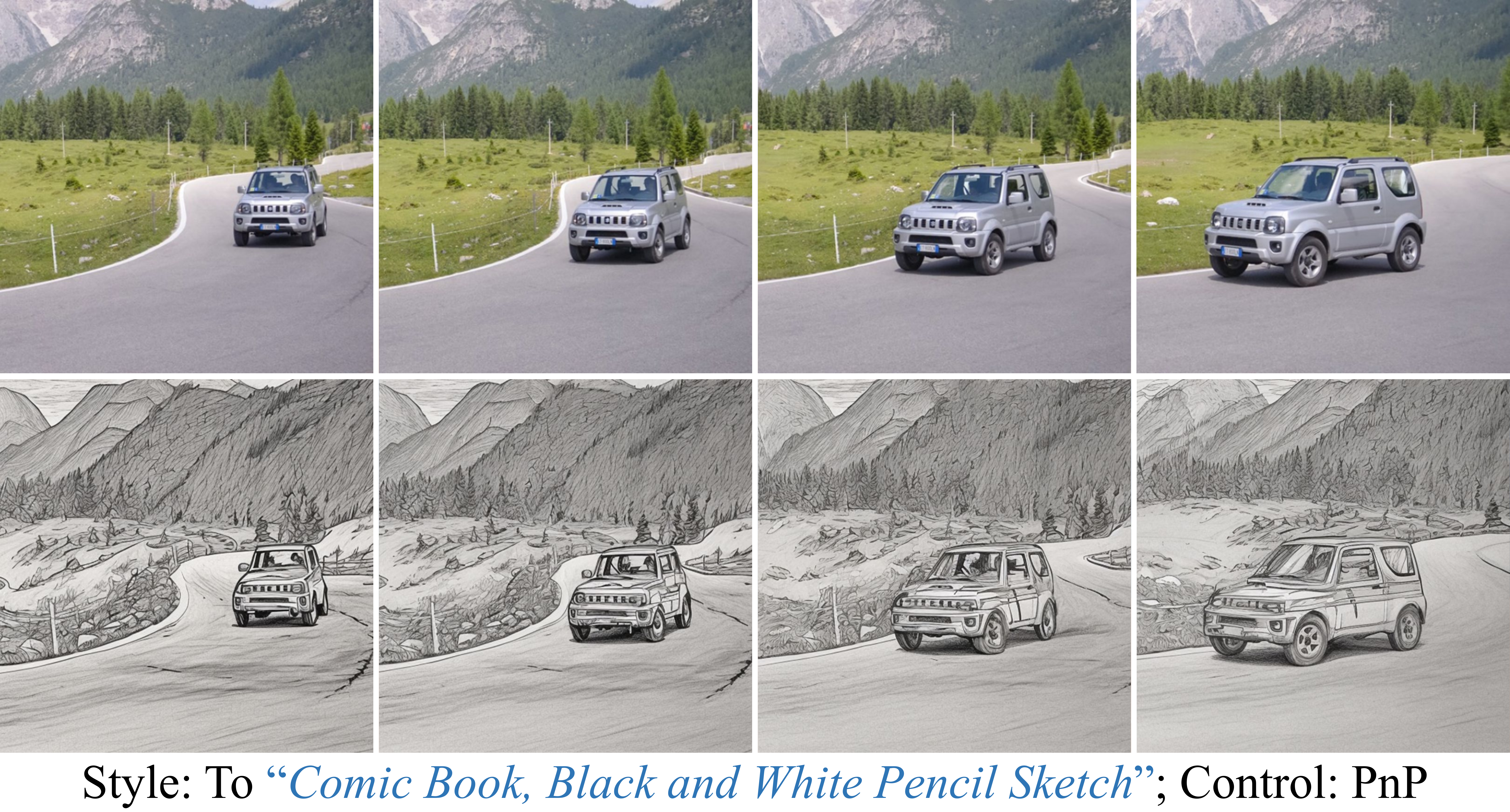} &   \includegraphics[width=\linewidth]{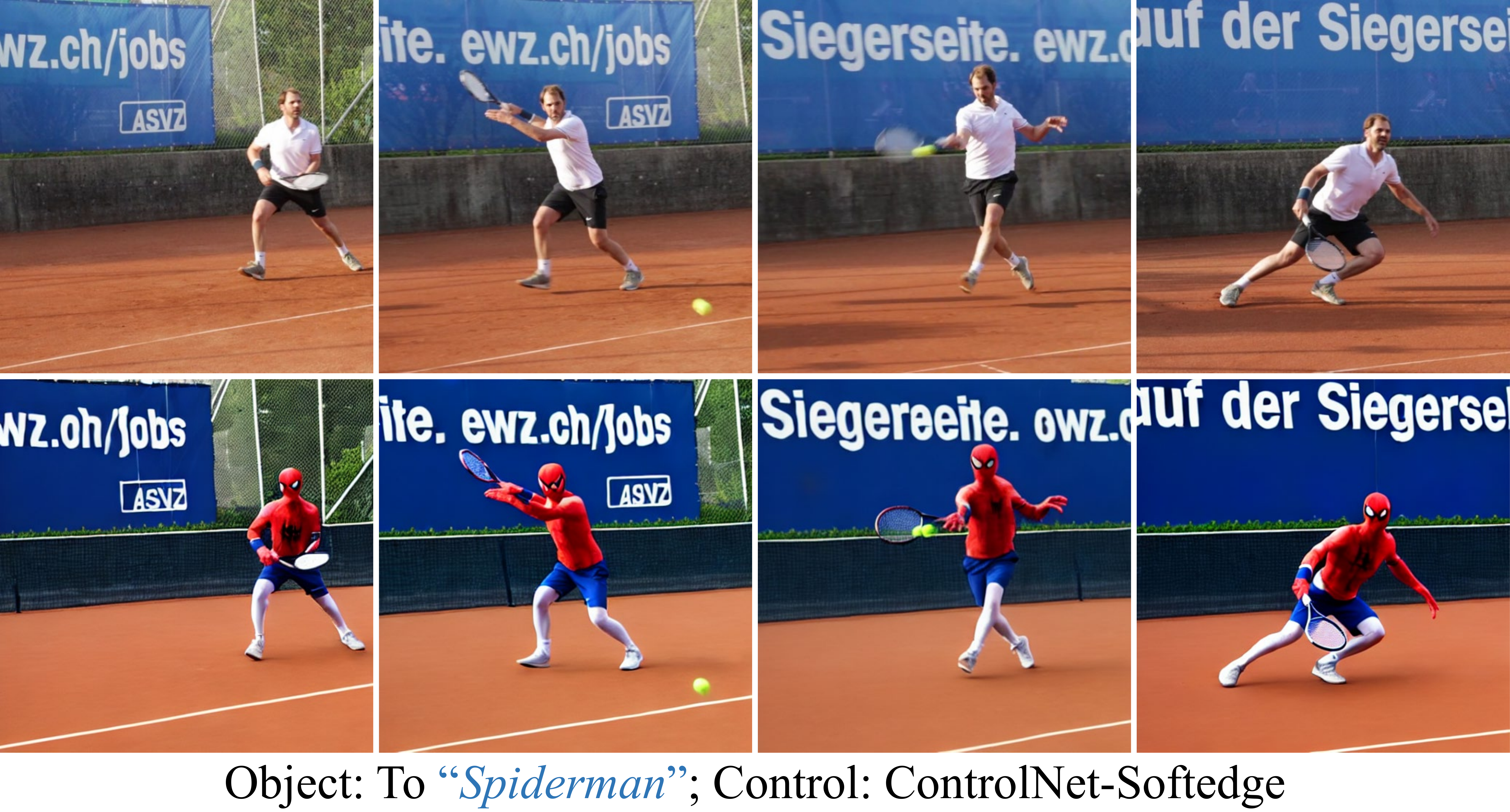} \\
  \includegraphics[width=\linewidth]{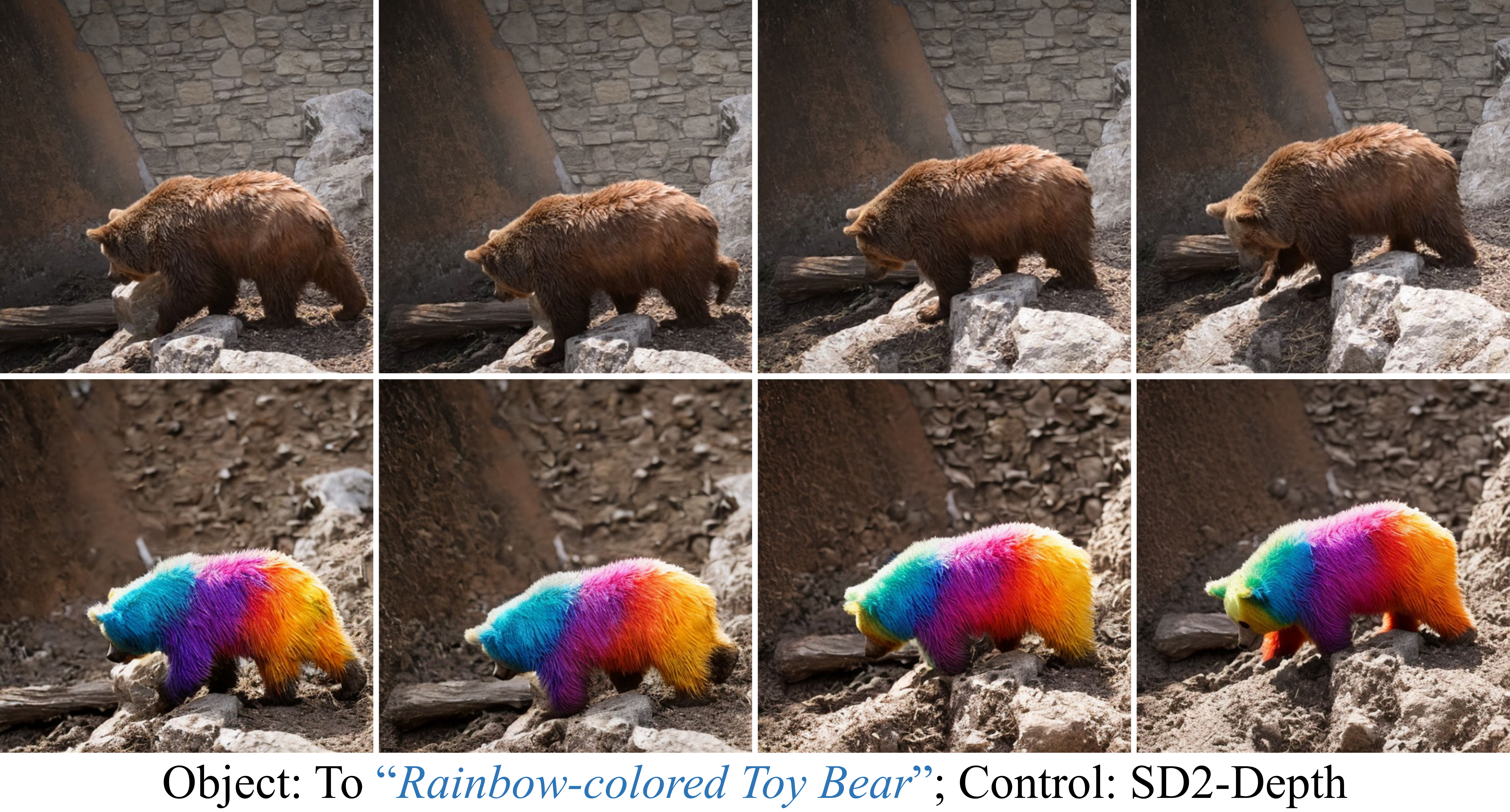} &  
 \includegraphics[width=\linewidth]{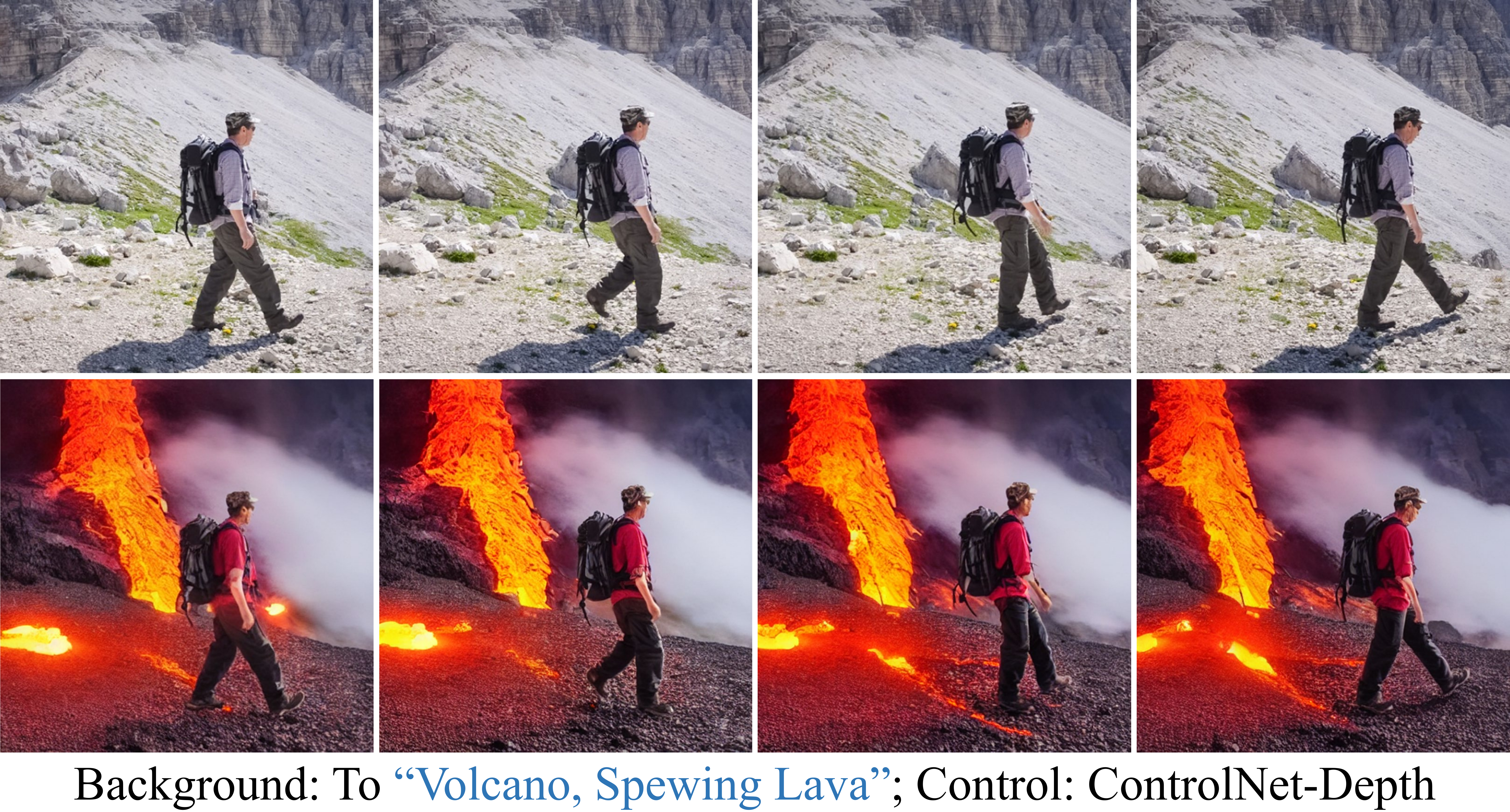}
\end{tabular}
\vspace{-0.2cm}
\caption{Sample editing results of our method. Our method can seamlessly integrate with existing controlling methods~\cite{tumanyan2023plug,zhang2023adding,rombach2022high} to temporal consistently edit videos in various aspects. We label each sample with the edit prompt and the applied controlling method.}
\label{fig:samples}
\end{figure*}

\subsection{Experiment Setting}
Our method performs video editing with a pretrained text-to-image model and an existing image-controlling method.
%
In this work, we use Stable Diffusion (SD)~\cite{rombach2022high} (version 1.5) as the image generator, and 
DDIM scheduler~\cite{song2020denoising} with sampling step $T=50$ for inversion and sampling. 
For the controlling method, we combine our method with Plug-and-Play (PnP)~\cite{tumanyan2023plug}, ControlNet~\cite{zhang2023adding}, and SD2-Depth. 
The parameters are chunk length $B=4$ and merging ratio $p=0.9$ and $0.8$ for local and global merging. 
More results and videos are available in the supplementary material. All the source code and datasets will be released.

\noindent \textbf{Dataset.}
Similar to prior works~\cite{ceylan2023pix2video, qi2023fatezero,wu2023cvpr}, we select 20 videos from DAVIS~\cite{Perazzi2016} as source videos for performance evaluation.  
These videos include a range of motion from slow to fast and feature various subjects such as humans, vehicles, and animals. 
We edit each of the 20 videos with three types of prompts:
(i) Style prompts edit the global style. (ii) Object prompts edit the object's appearance and attributes. (iii) Background prompts change the video background.
To obtain edit prompts, we use some prompts from~\cite{wu2023cvpr} and generate the others using GPT-3.5~\cite{ray2023chatgpt}. 
%
We generate 60 edited videos for evaluation, each containing 32 frames with a resolution of ~$512\times 512$.

\noindent \textbf{Metrics.}
We note that existing video editing methods use different metrics to evaluate the editing performance. 
In our experiment, we incorporate metrics used by prior works~\cite{ceylan2023pix2video,qi2023fatezero} and new measures. 
We assess the editing performance based on three key criteria: 
(i) Temporal Consistency: This includes Interpolation Error and PSNR~\cite{jiang2018super}, Warp Error, and Frame CLIP Score. 
(ii) Text Alignment: This includes Directional CLIP Score~\cite{gal2022stylegan} and Text CLIP Score. 
(iii) User Study.
Based on the Interpolation Error and PSNR used to evaluate video interpolation performance in previous studies~\cite{jiang2018super}, we propose to measure the video continuity by interpolating a target frame by its previous and next frames and computing the root-mean-squared (RMS) difference and PSNR between the estimated and target frames. The metric better reflects the generated video continuity itself without relying on the source video optical flow. 
We use the Directional CLIP Score~\cite{gal2022stylegan} to evaluate the consistency between image and prompt change. It computes the cosine similarity in CLIP space between the difference in frames and the difference in prompts from source to edit.
For more details about the metrics, please refer to the supplementary material. 
%
We conduct a user study with 10 out of 60 edited videos. Users choose their preferred video from the results edited by both baselines and our method. The user preference rate is used as the final metric.

\noindent \textbf{Baselines.} 
We evaluate our method against four state-of-the-art video editing techniques: Text2Video-Zero~\cite{khachatryan2023text2video}, Tune-A-Video~\cite{wu2023tune}, vid2vid-zero~\cite{wang2023zero}, and Pix2Video~\cite{ceylan2023pix2video}.
All the methods are implemented using default settings except for vid2vid-zero. We have to turn off its Spatial-Temporal attention that includes all frames in the self-attention, which is infeasible to fit the GPU memory (40GB) when processing 32-frame videos.
Text2Video-Zero allows zero-shot text-to-video generation, and we apply it with depth control to perform video editing.
Tune-A-Video finetunes the model on the source video frames before sampling the edited video.
We use StableDiffusion v1.5 for the first three methods and StableDiffusion2-Depth~\cite{rombach2022high} for Pix2Video as it requires a depth-conditioned model by default.
It is worth noting that all the baseline methods use some self-attention extension that includes multiple frames.


\subsection{Main Results}
\label{sec:results}
\begin{figure*}[t]
    \centering
    \includegraphics[width=.9\linewidth]{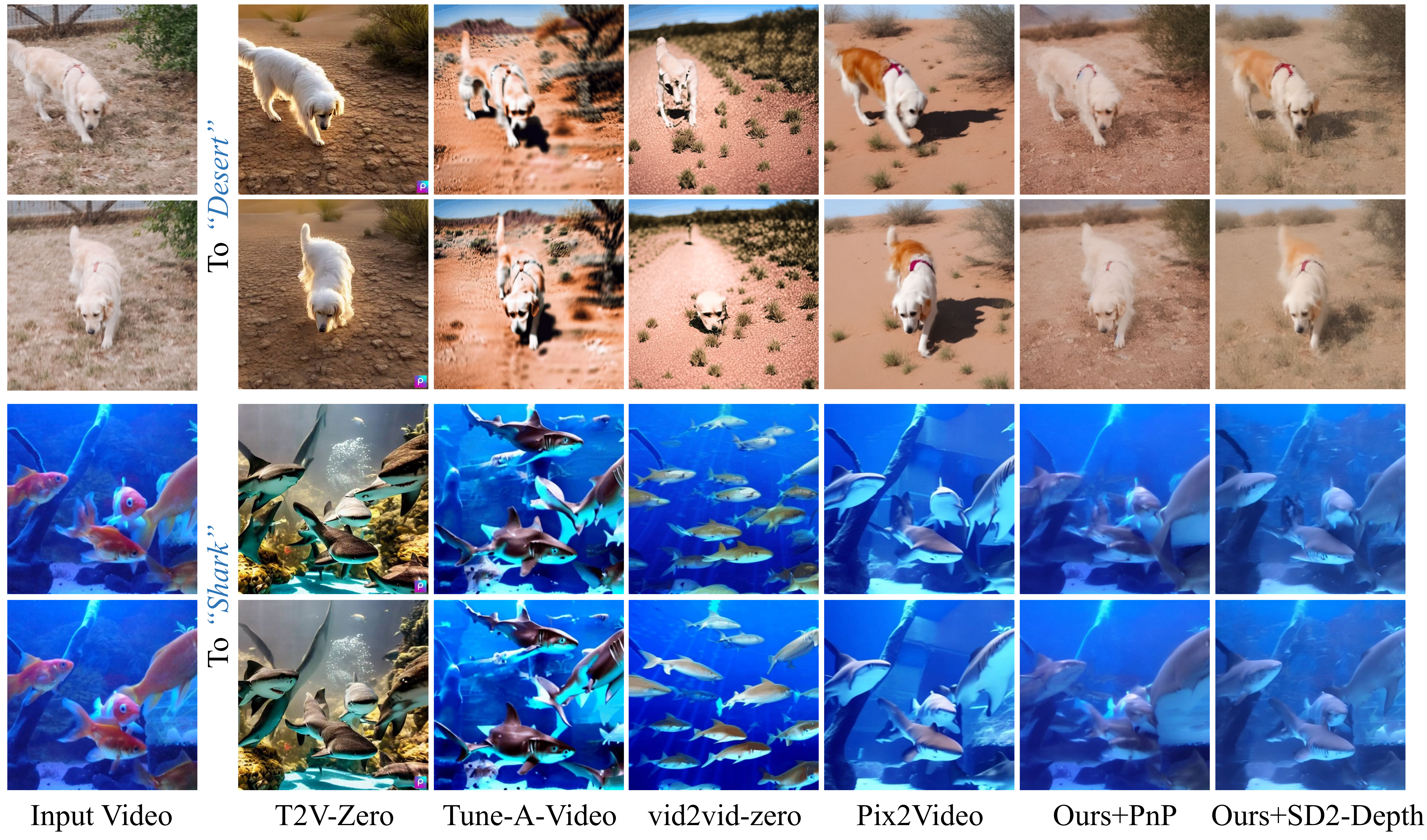}
    \caption{Qualitative comparison of our method and baselines. The editing results of our method are consistent over time in global style and local texture and preserve the source frame structure well.}
    \label{fig:qual_comp}
\end{figure*}
\begin{table*}[t]\scriptsize
    \centering
    \begin{tabular}{lccccccc}
    \toprule
        &  \multicolumn{4}{c}{Temporal Consistency}&  \multicolumn{2}{c}{Text Alignment}& User Study\\ 
    \cmidrule(lr){2-5}\cmidrule(lr){6-7}\cmidrule(lr){8-8}
 &
 Inp. Err.$\downarrow$ &
 Inp. PSNR$\uparrow$
 & Warp Err.$\downarrow$& Frame C.s.$\uparrow$& Directional C.s.$\uparrow$&  Text C.s.$\uparrow$&Preference Rate$\uparrow$\\ 
    \midrule
         Text2Video-Zero~\cite{khachatryan2023text2video}&  0.203&  19.4&  0.094&  0.969&  0.139&  0.282& 0.167\\ 
         Tune-A-Video~\cite{wu2023tune}&  0.235&  18.1&  0.068&  0.957&  0.132&  0.273& 0.089\\ 
         vid2vid-zero~\cite{wang2023zero}&  0.219&  18.6&  0.049&  0.957&  0.118&  0.270& 0.030\\
 Pix2Video*~\cite{ceylan2023pix2video}& 0.139& 22.5& 0.020& 0.968& \textbf{\textcolor{blue}{0.166}}& \textbf{\textcolor{red}{0.285}}&\textbf{\textcolor{blue}{0.170}}\\ 
    \midrule
 Ours + ControlNet-Depth~\cite{zhang2023adding}& 0.154& 22.1& 0.026& \textbf{\textcolor{blue}{0.973}}& 0.159& \textbf{\textcolor{blue}{0.284}} &
 \multirow{2}{*}{\textbf{\textcolor{red}{0.544}}}\\ 
 Ours + PnP~\cite{tumanyan2023plug}& \textbf{\textcolor{blue}{0.111}}& \textbf{\textcolor{blue}{25.0}}& \textbf{\textcolor{blue}{0.013}}& \textbf{\textcolor{red}{0.975}}& 0.140& 0.271&
\\ 
 Ours + SD2-Depth~\cite{rombach2022high}*& \textbf{\textcolor{red}{0.105}}& \textbf{\textcolor{red}{25.6}}& \textbf{\textcolor{red}{0.012}}& 0.971& \textbf{\textcolor{red}{0.168}}& 0.282&-\\
    \bottomrule
    \end{tabular}
    \caption{Quantitative evaluation results. Red and blue indicates the best and second-best result. *: Use the same base model as Pix2Video, SD2-Depth~\cite{rombach2022high}. Others use SDv1.5. C.s.: CLIP Score. Inp. Err.: Interpolation Error.}
    \label{tab:full}
\end{table*}

\noindent \textbf{Qualitative Evaluation.}
Fig.~\ref{fig:qual_comp} compares our editing results with baseline methods on evaluation videos.
%
While Text2Video-Zero~\cite{khachatryan2023text2video} produces high-quality frames, it lacks continuity between them.
The edited frames by Text2Video-Zero do not align with the source frame in appearance due to the random initial noise it used.
Tune-A-Video~\cite{wu2023tune} struggles to learn the motion of the source video and fails when the motion is complex.
The edited frames by Tune-A-Video contain wave-like jittering, degrading the video quality.
vid2vid-zero~\cite{wang2023zero} generates unstable video and fails to preserve the frame structure.
Pix2Video~\cite{ceylan2023pix2video} achieves good consistency between edited frames but generates unnatural jittering and blurring results.
In contrast, VidToMe generates consistent frames that adhere to the edit prompt while preserving the source frame structure. Fig.~\ref{fig:samples} showcases more sample editing results. Our method can integrate seamlessly with existing controlling methods, providing users with more control over the editing process and enabling video editing in various aspects.

\noindent\textbf{Quantitative Evaluation.}
We present quantitative evaluation results in Table~\ref{tab:full}.
The first three baselines do not perform well in terms of temporal consistency.
Text2Video-Zero~\cite{khachatryan2023text2video} uses random noise instead of noise inverted from source frames, which results in frames that are different from the source frames, as shown in the qualitative study.
Though some subjects appreciate the diversity in its results, its continuity is unsatisfactory.
Tune-A-Video~\cite{wu2023tune} and vid2vid-zero~\cite{wang2023zero} are not preferred by subjects due to temporal inconsistency.
Pix2Video~\cite{ceylan2023pix2video} achieves higher temporal consistency than the other baselines. 
It uses SD2~\cite{rombach2022high}, which has better generation fidelity than SDv1.5 and thus gets a high directional and text CLIP Score. However, its results still suffer from jittering and lack of long-term consistency.
Our proposed VidToMe achieves better temporal consistency and text alignment than the first three baselines. It also outperforms Pix2Video regarding temporal consistency when using the same base model, SD2-Depth.
Furthermore, Our editing results are preferred by over half of the subjects.
Our method offers flexibility in balancing consistency and alignment by using different control methods such as ControlNet-Depth~\cite{zhang2023adding} or PnP~\cite{tumanyan2023plug}.

\subsection{Ablation Studies}

\noindent \textbf{Efficiency Analysis.}
\begin{figure}[t]
    \centering
    \includegraphics[width=.8\linewidth]{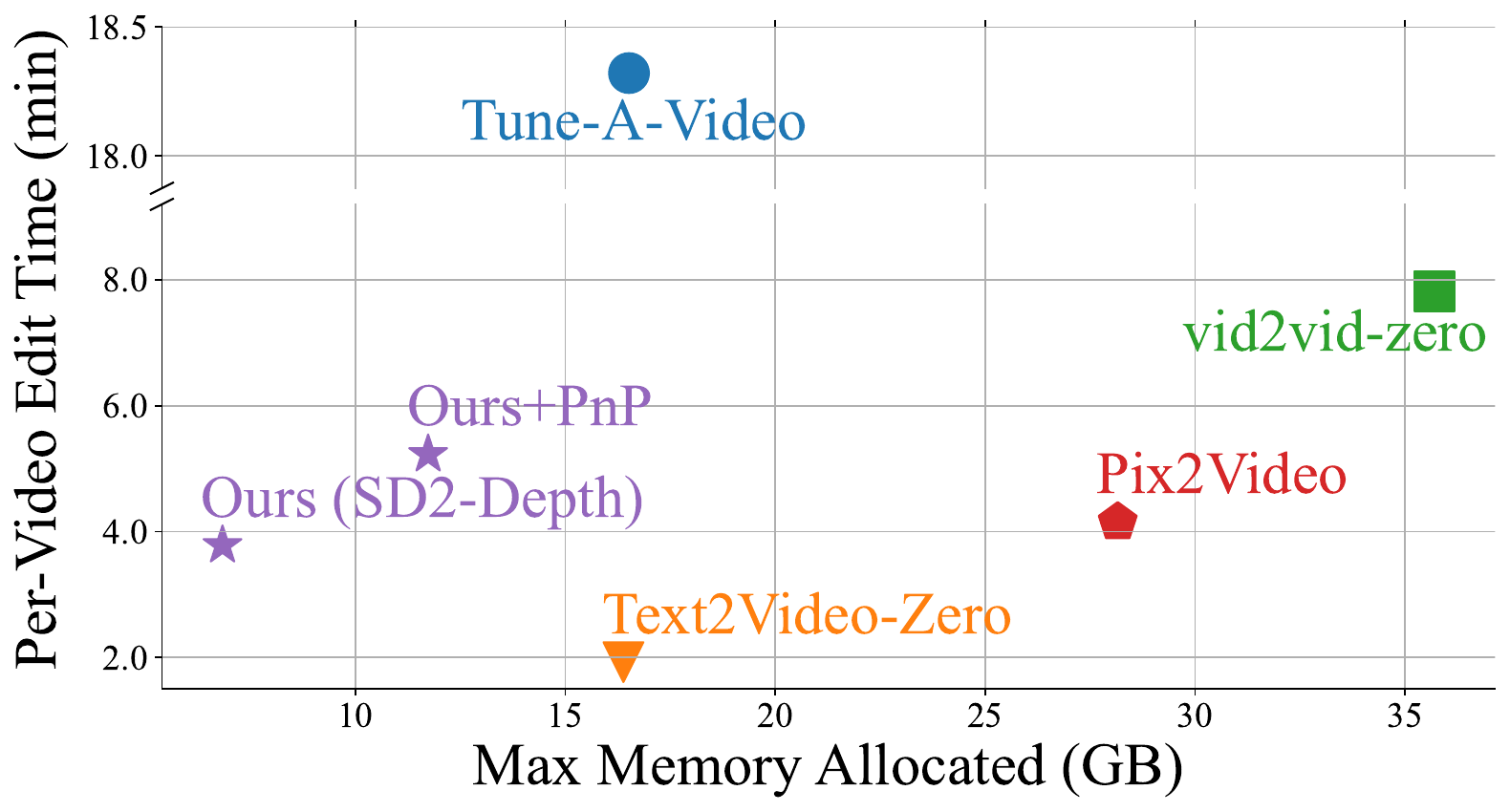}
    \caption{Editing efficiency comparison. We consider both time (time to perform one editing) and space (max GPU memory allocated in editing) efficiency. Test on one NVIDIA Tesla A100.}
    \label{fig:eff}
\end{figure}
In Fig.~\ref{fig:eff}, we compare the editing efficiency of different methods with their default setting. Most other methods require either high memory capacity or a long editing time. Although Text2Video-Zero~\cite{khachatryan2023text2video} edits videos quickly without noise inversion, its performance is poor. On the other hand, our proposed method, VidToMe, reduces memory consumption with video token merging while generating videos quickly. With a minimal memory consumption of less than 7 GB, our method can run on some personal devices.

\noindent \textbf{Multi-frame Self-Attention.}
\begin{table}\scriptsize
    \centering
    \begin{tabular}{lcc}
    \toprule
         &  Inp. Err. $\downarrow$&  Directional C.s. $\uparrow$\\
    \midrule
         Per-frame S.A.& 0.253 & 0.165 \\
         Extended S.A.& 0.140 &  0.168\\
 VidToMe S.A. (Ours)& 0.105 & 0.168\\
 \bottomrule
    \end{tabular}
    \caption{Performance comparison of different multi-frame self-attention (S.A.) operations.}
    \label{tab:sa}
\end{table}
We ablate on multi-frame self-attention choices in video editing. In per-frame self-attention, each frame is processed separately, leading to inconsistencies. On the other hand, extending self-attention to include multi-frame tokens is adopted by most existing methods. This method enables cross-frame attention and produces better consistency. Our approach merges tokens to enforce temporal consistency, achieving the best performance without sacrificing the editing effect.

\noindent \textbf{Token Merging Operation.}
\begin{figure}[t]
    \centering
    \includegraphics[width=0.8\linewidth]{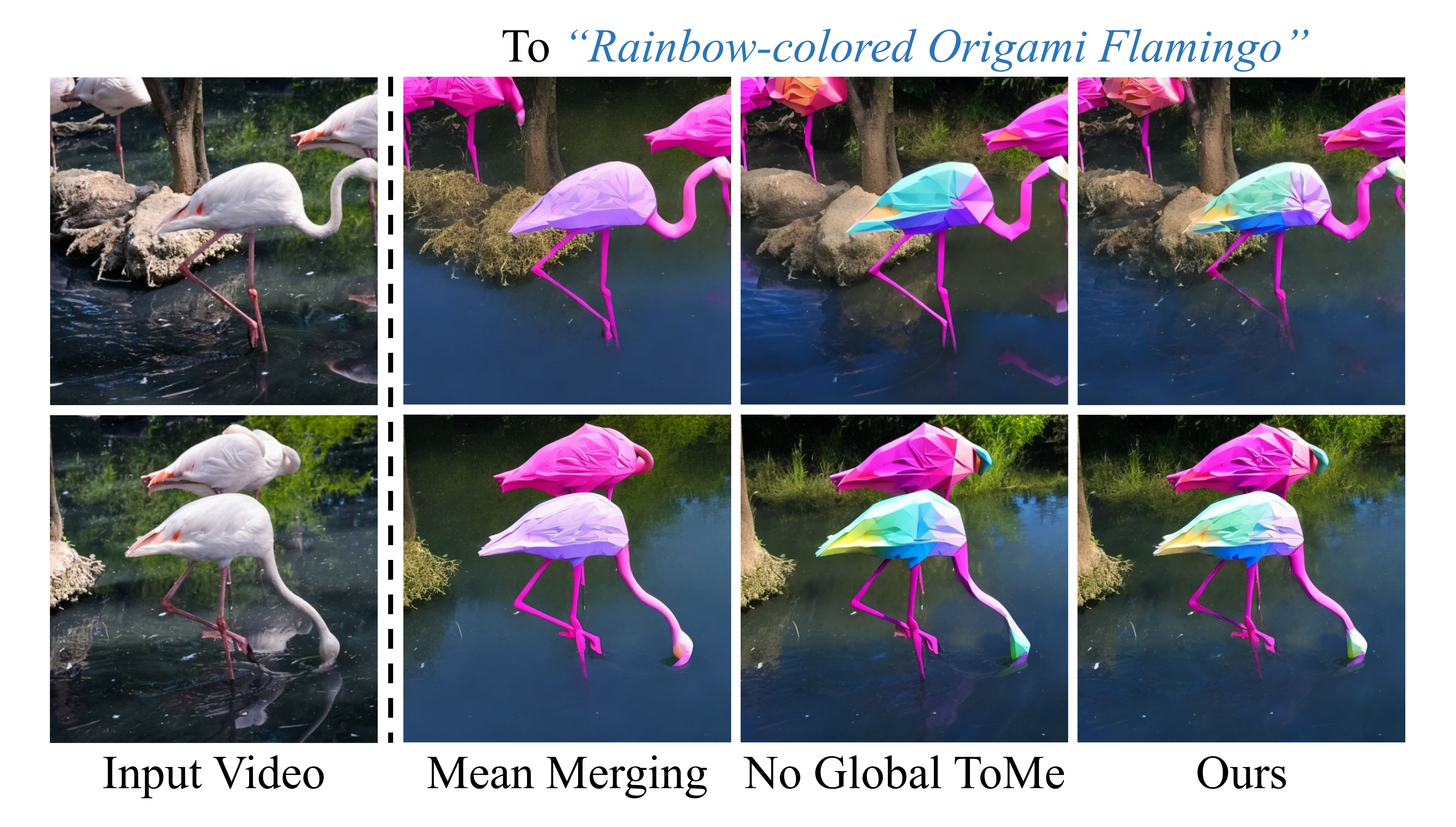}
    \caption{Ablation on Token Merging Operations. Merging tokens by mean instead of replacement reduces the edit fidelity. Without global token merging, the feather color changes.}
    \label{fig:abl-tome}
\end{figure}
We ablate on our token merging choice in Fig.~\ref{fig:abl-tome}. 
The original ToMe algorithm merges tokens by averaging their values. However, we find that this can lead to a lack of diversity and randomness in the generated videos, such as the flamingo features being single-colored. Therefore, we directly replace the value of the merged tokens with $dst$ tokens. Global token merging is crucial for keeping long-term consistency in videos. Without it, our method fails to maintain consistent rainbow feather colors throughout the video.

\noindent \textbf{Local Token Merging Strategy.}
\begin{figure}[t]
    \centering
    \includegraphics[width=\linewidth]{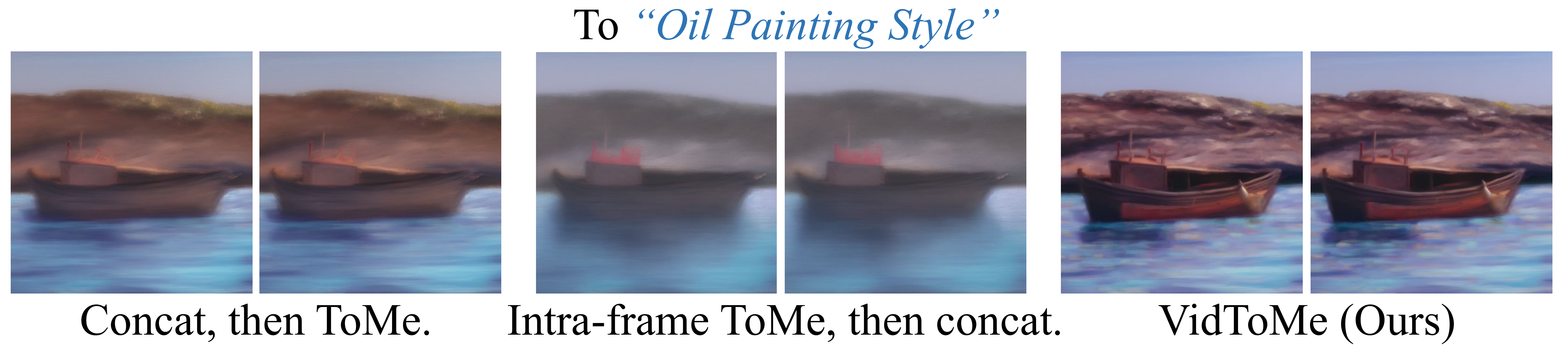}
    \caption{Ablation on local token merging strategy. Try: (i) Concatenate multi-frame tokens and then perform vanilla ToMe. (ii) Perform ToMe in each frame, then concatenate their tokens. (iii) Our local token merging strategy. We control the token number after merging to be the same.} 
    \label{fig:ablation-ltm}
\end{figure}
There are several possible strategies to merge tokens through multiple consecutive frames. One straightforward approach is to perform the original ToMe~\cite{bolya2023tomesd} among tokens from all frames, where $dst$ tokens are randomly selected. Or we can apply ToMe in each frame first and then concatenate all merged tokens. Fig.~\ref{fig:ablation-ltm} shows that these two methods result in blurry frames, while our local merging strategy preserves the quality of the generated frames.

\noindent \textbf{Limitations.}
Our method has two main limitations. First, the editing capability of our method depends on the performance of the selected image editing technique. If the editing technique fails on a single frame, our method also fails to edit the entire video. Second, although our similarity-based matching performs well in most cases, it has room for improvement. Objects with similar features are sometimes incorrectly merged and mixed in the output results. We plan to explore a more precise token-matching approach in the future to address these issues.
\section{Conclusion}
\label{sec:conclusion}

This work proposes a diffusion-based zero-shot video editing method, VidToMe.
Our approach unifies and compresses internal diffusion features by matching and merging tokens across video frames in the self-attention module during generation, resulting in temporally consistent edited video frames.
We implement VidToMe as lightweight token merging and unmerging blocks attached to the self-attention module, making it compatible with any existing image editing method and diffusion models.


{
    \small
    \bibliographystyle{ieeenat_fullname}
    \bibliography{main}
}

\end{document}